\documentclass[10pt,twocolumn,letterpaper]{article}

\usepackage{iccv}
\usepackage{times}
\usepackage{epsfig}
\usepackage{graphicx}
\usepackage{amsmath}
\usepackage{amssymb}
\usepackage{multirow}
\usepackage{booktabs}
\usepackage{caption}

\usepackage[pagebackref=true,breaklinks=true,letterpaper=true,colorlinks,bookmarks=false]{hyperref}

\iccvfinalcopy 

\def\iccvPaperID{8865} 

\ificcvfinal\fi

\begin{document}

\title{Pose-Free Neural Radiance Fields via Implicit Pose Regularization}

\author{Jiahui Zhang$^{1}$
\quad Fangneng Zhan$^{2}$
\quad Yingchen Yu$^{1}$
\quad Kunhao Liu$^{1}$\\
\quad Rongliang Wu$^{1}$
\quad Xiaoqin Zhang$^{3}$
\quad Ling Shao$^{4}$
\quad Shijian Lu \thanks{Corresponding author, E-mail: shijian.lu@ntu.edu.sg}\ $^{1}$\\
\\[1mm]
{\small $^1$Nanyang Technological University\quad$^2$Max Planck Institute for Informatics}\\[0.1mm]
{\small $^3$Wenzhou University\quad$^4$UCAS-Terminus AI Lab, UCAS}
}
\maketitle

\ificcvfinal\thispagestyle{empty}\fi



\begin{figure*}[t]
\centering
\includegraphics[width=1.\textwidth]{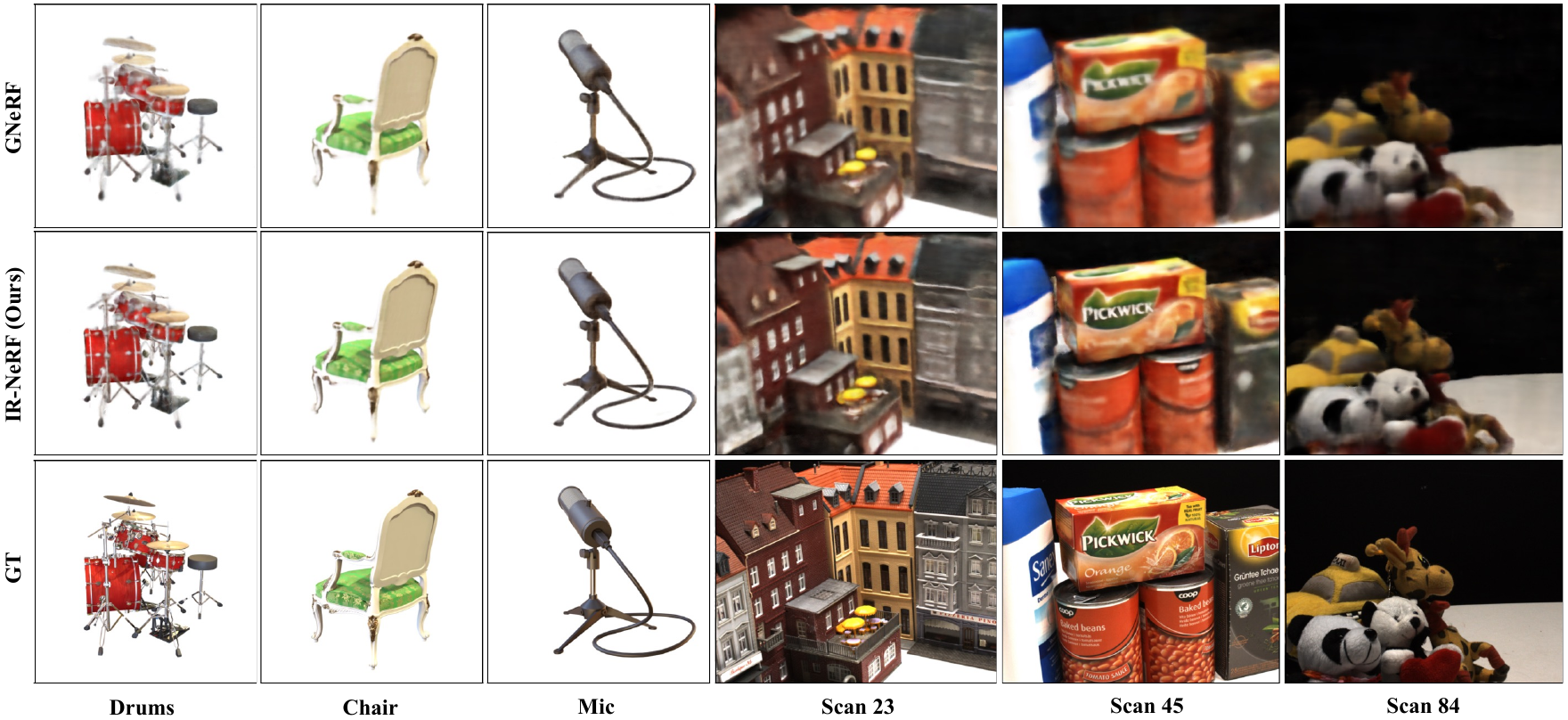}
\caption{
\textbf{Examples of novel view synthesis by GNeRF and our IR-NeRF.} The samples are from Synthetic-NeRF \cite{mildenhall2020nerf} and DTU \cite{jensen2014large}. It can be observed that the IR-NeRF synthesized novel views have less artifacts and finer details than GNeRF. 
}
\label{teaser}
\end{figure*}

\begin{abstract}
   Pose-free neural radiance fields (NeRF) aim to train NeRF with unposed multi-view images and it has achieved very impressive success in recent years. 
   Most existing works share the pipeline of training a coarse pose estimator with rendered images at first, followed by a joint optimization of estimated poses and neural radiance field.
   However, as the pose estimator is trained with only rendered images, the pose estimation is usually biased or inaccurate for real images due to the domain gap between real images and rendered images, leading to poor robustness for the pose estimation of real images and further local minima in joint optimization.
   We design \textit{IR-NeRF}, an innovative pose-free NeRF that introduces implicit pose regularization to refine pose estimator with unposed real images and improve the robustness of the pose estimation for real images.
   With a collection of 2D images of a specific scene, IR-NeRF constructs a scene codebook that stores scene features and captures the scene-specific pose distribution implicitly as priors. Thus, the robustness of pose estimation can be promoted with the scene priors according to the rationale that a 2D real image can be well reconstructed from the scene codebook only when its estimated pose lies within the pose distribution. Extensive experiments show that IR-NeRF achieves superior novel view synthesis and outperforms the state-of-the-art consistently across multiple synthetic and real datasets. 
\end{abstract}

\section{Introduction}

Novel view synthesis has recently achieved remarkable progress, largely driven by the development of neural radiance fields (NeRF) \cite{mildenhall2020nerf} that learns 3D scene representations from multi-view 2D images and can generate novel views with superb multi-view consistency. However, most existing works rely heavily on accurate camera poses of the multi-view 2D images which are complicated to collect and not available in many existing image datasets. The camera pose constraint can be mitigated by leveraging structure-from-motion (SfM)~\cite{hartley2003multiple, schonberger2016structure} that allows estimating camera poses from multi-view 2D images. On the other hand, SfM requires keypoint detection and is prone to errors while handling objects and scenes with low texture or repeated visual patterns. How to train effective NeRF with unposed multi-view images has become one bottleneck for the wide adoption of NeRF in various 3D synthesis tasks.

Several studies attempt pose-free NeRF by training NeRF with unposed multi-view images. One approach is to train NeRF with certain inaccurate camera poses or prior knowledge about camera pose distributions. For example, \cite{wang2021nerf} jointly optimizes NeRF and camera poses to alleviate the requirement for accurate camera poses. BARF~\cite{lin2021barf} exploits bundle adjusting to train NeRF with imperfect camera poses. \cite{chng2022garf} introduces Gaussian activated radiance field that employs Gaussian activation to avoid falling into local minima. Nevertheless, this approach still requires reasonable camera pose initialization that is often not easy to obtain. Another approach does not require any pose information in training. For example, GNeRF~\cite{meng2021gnerf} first trains coarse NeRF with randomly initialized camera poses and predicts coarse camera poses, and then jointly refines them with the NeRF training process. However, the pose estimator in GNeRF is trained only with images rendered by the coarse NeRF. The pose prediction for real images is biased or inaccurate due to the domain gap between rendered images and real images, leading to poor robustness of pose estimation for real images and local minima \cite{meng2021gnerf} while jointly refining NeRF and camera poses.


We propose IR-NeRF, an innovative pose-free NeRF that introduces Implicit Regularization to promote the robustness of pose estimator for real images.
Specifically, given a set of multi-view images of a scene, a scene codebook is first constructed which stores the scene features and encodes scene-specific pose distribution implicitly as priors. 
With that, a pose-guided view reconstruction scheme is then designed to refine the pose estimator with unposed real images, based on the rationale that a real image can be reconstructed well from the codebook only when its estimated camera pose lies within the scene-specific pose distribution. 
With the accurate camera poses predicted by the refined pose estimator,
IR-NeRF can jointly optimize NeRF and estimated camera poses  without getting stuck in local minima, yielding accurate NeRF representations with superior novel view synthesis as illustrated in Fig.~\ref{teaser}.

The contributions of this work are threefold. 
\textit{First}, we propose IR-NeRF, a novel pose-free NeRF that introduces implicit pose regularization that enables effective NeRF training with unposed multi-view images. 
\textit{Second}, with a set of multi-view 2D images of a scene, we construct a scene codebook that encodes scene features and implicitly captures scene-specific camera pose distribution as priors.
\textit{Third}, we design a pose-guided view reconstruction scheme that utilizes the scene priors to refine pose estimator with unposed real images, which allows to promote the robustness of the pose estimator.


\section{Related Work}

\paragraph{Neural Radiance Fields.} NeRF~\cite{mildenhall2020nerf} encodes 3D locations and 2D viewing directions into RGB colour and volume density, and it has demonstrated very impressive performance in novel view synthesis. With implicit scene representation and differentiable volume rendering, NeRF has been developing quickly recently with a number of variants and extensions, including generative radiance fields ~\cite{schwarz2020graf, niemeyer2021giraffe, gu2021stylenerf}, generalizable radiance fields \cite{chen2021mvsnerf, xu2022point, xu2023wavenerf}, dynamic scene representations \cite{park2021nerfies, du2021neural, gao2021dynamic, guo2021ad, tretschk2021non}, fast scene representations \cite{muller2022instant, garbin2021fastnerf, reiser2021kilonerf}, neural surface representations \cite{wang2021neus, zhang2021ners} and unbounded scene representations \cite{zhang2020nerf++, barron2022mip}. 
However, most existing work requires accurate camera poses of 2D training images for proper NeRF training, whereas camera pose collection is often complicated and prone to errors which impairs the scalability of NeRF greatly. As a comparison, our proposed IR-NeRF can train effective NeRF with a set of unposed multi-view images.

\paragraph{Pose-Free NeRF} 
Pose-free NeRF has attracted increasing attention recently for training effective NeRF with unposed images. Most existing methods manage to estimate the camera pose of training images, and they can be broadly grouped into two categories depending on whether they involve learning in camera pose estimation. Most non-learning methods \cite{zhang2020nerf++, martin2021nerf} exploit conventional techniques such as Structured-from-Motion (SfM) \cite{hartley2003multiple, faugeras2001geometry, wu2013towards, schonberger2016structure}) for camera pose estimation. However, conventional methods often have limited robustness and accuracy. For example, SfM estimates camera poses from key-point correspondences across images which does not work well for scenes with very sparse textures or repeating visual patterns. 

Methods in the second category estimate camera poses via learning. One typical approach trains pose-free NeRF with certain roughly initialized camera poses. For example, \cite{wang2021nerf} jointly optimizes initialized camera poses and NeRF model. \cite{lin2021barf} exploits bundle adjusting for coarse-to-fine camera pose registration and joint optimization of camera poses and NeRF. \cite{chng2022garf} employs Gaussian activation for pose estimation and NeRF optimization. \cite{jeong2021self} refines the initialized camera poses via self-calibration. However, this approach often produces degraded NeRF models when the initialized camera pose does not have reasonable accuracy. Another approach \cite{meng2021gnerf, zhang2022vmrf} learns NeRF with randomly initialized camera poses. For example, GNeRF~\cite{meng2021gnerf} introduces a pose estimator to directly
estimate camera poses from images. However, the pose estimator is trained only with rendered images, leading to inaccurate or biased predictions on real images used in NeRF training due to the domain gap between rendered images and real images. The poor robustness of pose estimator for real images tends to result in local minima in NeRF training. Our IR-NeRF introduces implicit pose regularization to refine pose estimator training with unposed real images, which enhances the robustness of pose estimation for real images, leading to superior pose-free NeRF.

\paragraph{Visual Codebook} Standard visual codebook \cite{van2017neural} aims to learn a discrete and compressed image representation via vector quantization, and it has been widely explored in the computer vision community. For example, \cite{esser2021taming} constructs a rich visual codebook to achieve high-resolution image synthesis with transformer. \cite{gu2022vector} combines the visual codebook with diffusion model \cite{sohl2015deep, ho2020denoising, dhariwal2021diffusion} for text-to-image generation. \cite{yu2021vector} proposes multiple improvements over
vanilla VQGAN \cite{esser2021taming} for improving  vector quantized image modeling tasks. In IR-NeRF, we design a novel scene codebook construction technique that adopts linear combination instead of vector quantization for implicit pose regularization. To the best of our knowledge, IR-NeRF is the first work that adapts visual codebook for pose-free NeRF optimization.

\section{Preliminary}



\paragraph{Camera Pose Estimation} Camera pose distribution is determined by camera poses of multi-view images of a specific 3D scene \cite{meng2021gnerf}. Specifically, camera positions are distributed on the surface of partial sphere which is determined by the radius of sphere and camera elevation range and camera azimuth range. Camera rotation depends on camera position, camera lookat points and camera lookup vector. As greater viewpoint uncertainty tends to lead to local minima while jointly optimizing camera poses and NeRF model \cite{meng2021gnerf}, it is critical to ensure that estimated camera poses are located within scene-specific camera pose distribution.


\paragraph{Neural Radiance Field} NeRF \cite{mildenhall2020nerf} is proposed to represent a 3D scene as a 5D function that is parameterized with MLP. It takes a 3D location $x \in \mathbb{R}^3$ and a 2D viewing direction $d \in \mathbb{S}^2$ as input and generates a RGB color $[r, g, b]$ and volume density $\sigma$ for this location. This process can be formulated by $F_\Theta: (x, d) \rightarrow ([r, g, b], \sigma)$, where $F$ and $\Theta$ denote MLP network and its parameters, respectively. Volume rendering is then adopted to render 2D images from NeRF scene representation by the accumulation of colors and densities at camera rays. To ensure the differentiability of the volume rendering, numerical quadrature is adopted to approximate the continuous integral by stratified sampling from depth bounds. Additionally, NeRF models are optimized by a photometric loss between the real and corresponding rendered pixel colors, which is formulated by sum of squared differences.

\section{Proposed Method}

\subsection{Overall Framework}
\begin{figure*}[t]
\begin{center}
\includegraphics[width=1\linewidth]{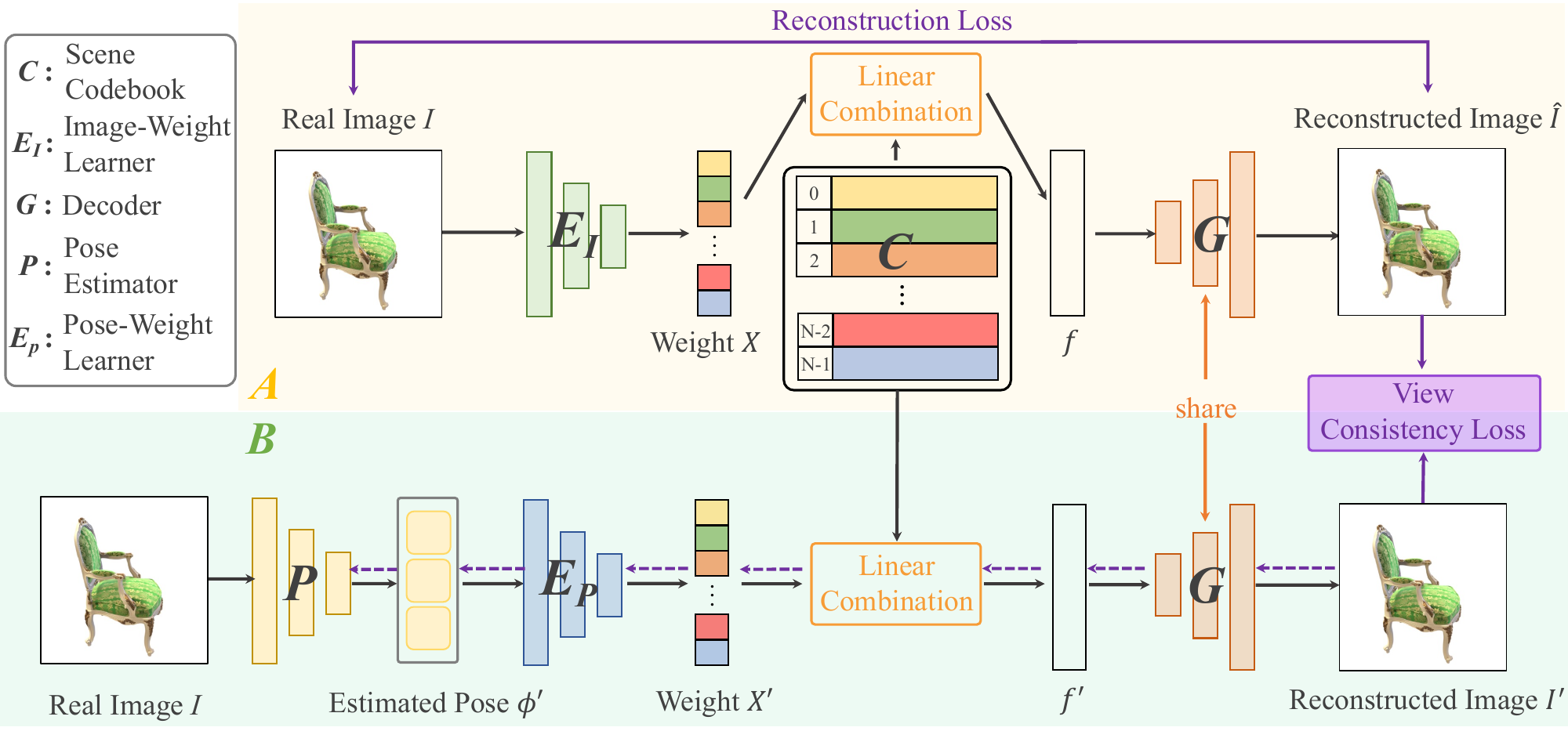}
\end{center}
\caption{
\textbf{Overview of the proposed implicit pose regularization.}
Part `A' in yellow and part `B' in green represent scene codebook construction and pose-guided view reconstruction, respectively. Leveraging image-weight learner $E_I$, scene codebook $C$ and decoder $G$, the real image $I$ can be reconstructed from a feature embedding $f$ which is constructed by linear combination of feature embeddings in the codebook. $E_I$, $C$ and $G$ are trained simultaneously via the image reconstruction process. Pose estimator $P$ predicts the camera pose $\phi'$ of the real image $I$ in training dataset. With the learned $C$ and $G$, image $I'$ corresponding to $\phi'$ is reconstructed by linear combination of learned feature embeddings in $C$, where combination weights $X'$ are derived from $\phi'$ through a pose-weight learner $E_P$. 
The view consistency loss between $I'$ and $\hat{I}$ regularizes the pose estimation.
The purple dashed line highlights the regularization process for pose estimation.
}
\label{model}
\end{figure*}

With initial camera poses $\Phi = \{\phi_i, i \in [1, T]\}$ randomly sampled from a predefined pose distribution following the settings in GNeRF \cite{meng2021gnerf}, the proposed IR-NeRF first learns a coarse NeRF with an adversarial loss, and then utilizes the trained NeRF to render images with $\Phi$. 
A pose estimator $P$ is trained in two steps to predict camera poses. First, it is trained by regressing initialized camera poses with rendered images as in \cite{meng2021gnerf}. Second, IR-NeRF introduces an implicit pose regularization to refine the pose estimator with unposed real images.
The implicit pose regularization for real images leads to robust pose estimation, 
as pose estimator trained with only rendered images is inaccurate for real images due to the domain gap between real images and rendered images.

As shown in Fig.\ref{model}, the key components in the implicit pose regularization are scene codebook construction and pose-guided view reconstruction with view consistency loss. 
Specifically, a scene codebook $C$ with scene features and scene-specific pose distribution is first learned by the reconstruction of the unposed real images in training dataset $\mathcal{I}$ used in NeRF training. 
Then, given real image $I$, pose-guided view reconstruction exploits pose estimator $P$ to predict camera pose $\phi'$ of image $I$, and further utilizes $\phi'$ to guide linear combination of feature embeddings in the learned scene codebook to reconstruct the corresponding image $I'$. 
As the trained $C$ and $G$ ensure that an image can be reconstructed well only when its estimated camera pose lies within accurate pose distribution, implicit pose regularization can be achieved with a view consistency loss $\mathcal{L}_c$ between $I'$ and $\hat{I}$.
We also jointly refine the learned coarse NeRF and predicted camera poses. With the implicit pose regularization, the joint refinement can effectively avoid falling into local minima.
Details of the designed scene codebook construction and pose-guided view reconstruction will be discussed in the ensuing section \ref{scc} and section \ref{pvr}, respectively.

\subsection{Scene Codebook Construction}
\label{scc}
The scene codebook construction allows to learn scene-specific pose distribution implicitly as priors which lays a base for the subsequent pose-guided view reconstruction. 
Instead of naively encoding input images to latent representations which fails to capture overall pose distribution, we design a novel scene codebook construction scheme with a linear combination which can serve as implicit distribution prior to achieve robust pose estimation.

As shown in Fig.~\ref{model}, the scene codebook construction consists of an image-weight learner $E_I$, a scene codebook $C = \{c_n\}_{n=1}^N \in \mathbb{R}^{N \times D}$ and a decoder $G$. The scene codebook is learned via the reconstruction of unposed real images. 
The image-weight learner $E_I$ is utilized to yield a collection of combination weights $X = \{x_n\}_{n=1}^N \in \mathbb{R}^N$ based on the real image $I$:
\begin{equation}
X = Softmax(E_I(I)), \quad x_n = \frac{e^{l_n}}{\sum_{j=1}^{N} e^{l_j}},
\end{equation}
where $Softmax(\cdot)$ denotes the softmax function and $l_n$ represents the Logits output of $E_I$ (before Softmax). The feature embedding $f$ of the real image $I$ is then constructed by linear combination of feature embeddings in codebook, which can be formulated as follows:
\begin{equation}
f = \sum_{n=1}^{N} c_n x_n
\end{equation}

With the feature embedding $f$, the real image $I$ can be reconstructed via the decoder $G$ 
by: 
\begin{equation}
    I \approx \hat{I} = G(f),
\end{equation}
where $\hat{I}$ denotes the reconstructed image. With an image reconstruction loss $\mathcal{L}_{rec}$, the scene codebook can be learned with the image-weight learner $E_I$ and the decoder $G$:
\begin{equation}
    \mathcal{L}_{rec}(E_I, C, G) = \| I - \hat{I} \|^2,
\end{equation}

 To reduce the difficulty of joint training of $E_I$, $C$ and $G$ and improve the training stability, we employ the pre-trained VGG19 \cite{simonyan2014very} to initialize the scene codebook $C$ by encoding a set of real images $[I_0, I_1, ... , I_T]$, where $T$ represents the number of real images. This process can be formulated as follows:
\begin{equation}
    C_{ini} = VGG([I_0, I_1, ... , I_T]),
\end{equation}
where $VGG(\cdot)$ represents the VGG19 network, $C_{ini}$ denotes the initialized scene codebook, which will be further optimized by image reconstruction loss $\mathcal{L}_{rec}$.

\subsection{Pose-Guided View Reconstruction} 
\label{pvr}

With the learned scene codebook $C$ and decoder $G$, it can be guaranteed that only images with camera poses within scene-specific pose distribution can be well-reconstructed. Under this rationale, we design pose-guided view reconstruction with view consistency loss to refine pose estimation with unposed real images. Based on the estimated camera pose $\phi'$ of real image $I$, the image $I'$ corresponding to $\phi'$ is constructed by linear combination of the learned feature embeddings in scene codebook. Specifically, a pose-weight learner $E_P$ is first utilized to produce a set of combination weights $X' = \{x'_n\}_{n=1}^N \in \mathbb{R}^N$ based on the estimated camera pose $\phi'$, which can be formulated as follows:
\begin{equation}
    X' = Softmax(E_P(\phi')), \quad x'_n = \frac{e^{l'_n}}{\sum_{j=1}^{N} e^{l'_j}},
\end{equation}
where $l'_n$ represents the Logits output of $E_P$ (before Softmax). The construction of feature embedding $f'$ corresponding to $\phi'$ can then be represented as $f' = \sum_{n=1}^{N} c_n x'_n$, where $c_n$ and $x'_n$ denote the $n$-th feature embedding in scene codebook and $n$-th combination weight, respectively. Finally, the corresponding image $I'$ can be reconstructed via the frozen decoder $G$ which focuses on decoding the feature embedding generated by linear combination of feature embeddings in scene codebook.

Leveraging the image $\hat{I}$ reconstructed from the shared decoder $G$ as pseudo ground truth, a view consistency loss $\mathcal{L}_c$ between the reconstructed image $I'$ and the pseudo ground truth can be formulated as below:
\begin{equation}
\mathcal{L}_c(P, E_P) =  \frac{1}{i} \sum_{i=1}^{S} \left\| I'_i - \hat{I}_i \right\|_{2}^2.
\end{equation}
If the camera pose $\phi'$ estimated by $P$ deviates from scene-specific pose distribution, the corresponding view $I'$ reconstructed by the learned $C$ and $G$ will not be aligned with the pseudo ground truth $\hat{I}$.
Thus, the robustness of pose estimation can be promoted as the out-of-distribution pose estimation are suppressed.

\begin{figure*}[t]
\begin{center}
\includegraphics[width=1\textwidth]{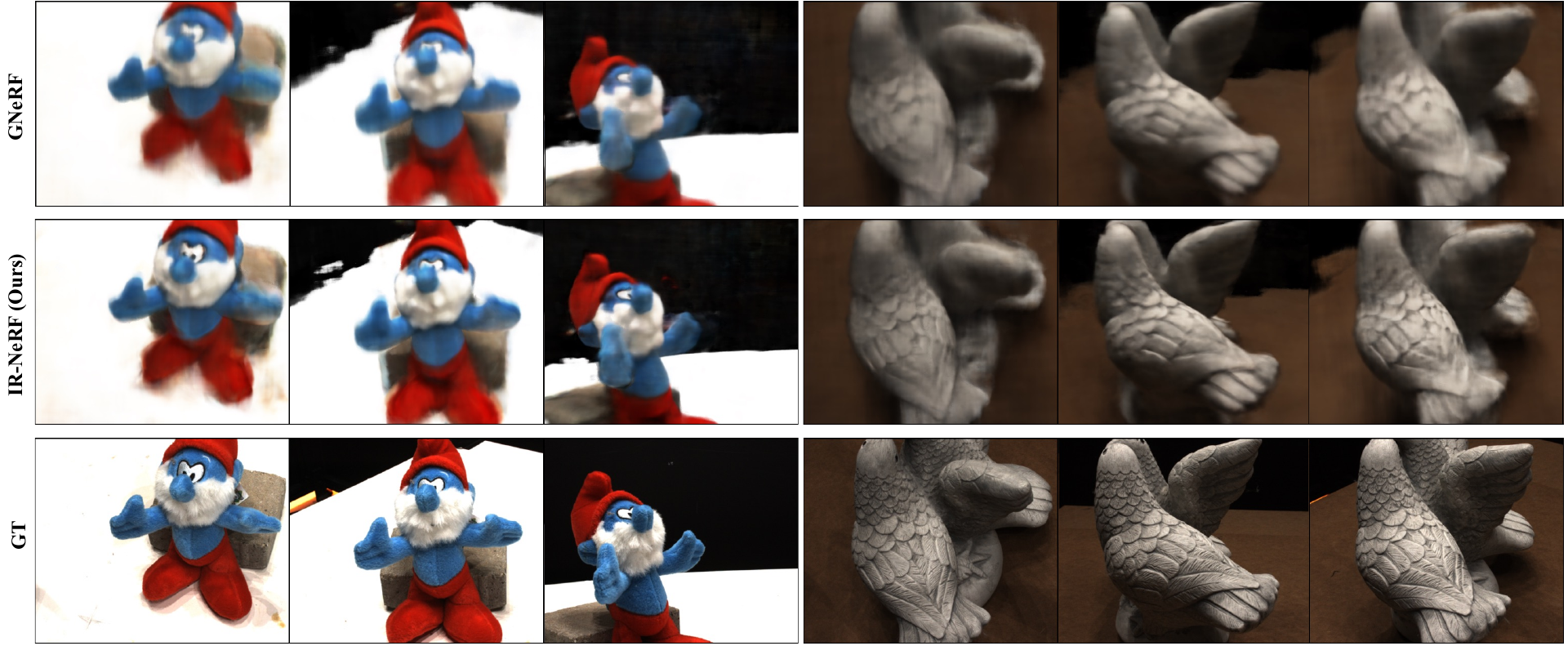}
\end{center}
\caption{
\textbf{Qualitative comparisons of IR-NeRF with GNeRF in novel view synthesis:} The comparisons are conducted over different views of scenes `scan82' and `scan109' in DTU, where `GT' denotes the ground-truth image. It is clear that IR-NeRF synthesizes high-fidelity images with less artifacts and finer details compared with GNeRF. \textbf{Zoom in for best view.}
}
\label{visual_compare}
\end{figure*}

\subsection{Training Process}

The training process of the proposed IR-NeRF includes coarse NeRF learning, camera pose estimation, and joint refinement of NeRF and predicted camera poses. For the coarse NeRF training, we introduce an adversarial loss to train a coarse NeRF $F$ with randomly initialized poses $\Phi$ due to lack of known camera poses. The adversarial loss $\mathcal{L}_{adv}$ can be defined as follows:
\begin{equation}
\begin{aligned}
    \mathcal{L}_{adv}(F, D) & =  \mathbb{E}_{I\sim P_{data}}[log(D(I))]\\
    & + \mathbb{E}_{F(\Phi)\sim P_{g}}[log(1 - D(F(\Phi)))], 
\end{aligned}
\end{equation}
where $D$ denotes the discriminator, $P_{data}$ and $P_g$ represent the distribution of images generated by NeRF and real images in training dataset, respectively.

For the camera pose estimation, we first employ MSE loss to optimize a coarse pose estimator $P$ with images rendered by the trained coarse NeRF as in GNeRF \cite{meng2021gnerf}. The pose estimator is then refined for real images via implicit pose regularization. Specifically, the scene codebook construction is performed with unposed real images under the supervision of the image reconstruction loss $\mathcal{L}_{rec}$. With the learned scene codebook and decoder, the pose estimator can be optimized to predict the camera poses of real images driven by the view consistency loss $\mathcal{L}_c$. With coarse NeRF and predicted camera poses, we also employ a photometric loss for joint optimization of the NeRF and camera poses. Specifically, we leverage the hybrid and iterative optimization scheme \cite{meng2021gnerf} for end-to-end training of the proposed IR-NeRF, where the pose estimation and joint optimization are interleaved in the training. Note that NeRF is frozen during camera pose estimation but is trainable during joint refinement.

\section{Experiment}

\subsection{Datasets and Implementation Details}

\paragraph{Datasets} 
\label{dataset}

Following GNeRF \cite{meng2021gnerf}, we conduct experiments on synthetic and real-world scenes with the same split of training and evaluation sets. For synthetic scenes, we use NeRF-Synthetic dataset \cite{mildenhall2020nerf} which consists of object-centric scenes with complex geometry. For each scene, we train with 100 multi-view training images which are resized to 400 by 400 pixels. The evaluation is conducted on eight images that are randomly selected from the test set. For real-world scenes, we employ six representative scenes in the DTU dataset~\cite{jensen2014large}. We randomly split the 49 images of each scene into training and test sets, where the training set includes 43 images of resolution $500 \times 400$ and the test set consists of the remaining 6 images. 

\paragraph{Implementation Details}

For predefined camera pose distribution, we follow the settings in GNeRF \cite{meng2021gnerf}. Specifically, the range of azimuth, elevation, sphere radius and camera lookat point are set at $[0^\circ, 360^\circ]$, $[0^\circ, 90^\circ]$, $4.0$ and $ (0,0,0)$, and $[0^\circ, 150^\circ]$, $[0^\circ, 80^\circ]$, $4.0$ and $\mathcal {N}(0, 0.01 ^2)$ for both synthetic and real-world datasets, respectively.
For camera poses, camera position and camera rotation are represented by a 3D embedding in Euclidean space and a continuous 6D embedding \cite{zhou2019continuity}, respectively. The camera pose embedding can be recovered to a transformation matrix by a Gram-Schmidt-like process \cite{zhou2019continuity}. For the architecture of IR-NeRF, the image-weight learner $E_I$, pose-weight learner $E_P$ and decoder $G$ are CNN-based, MLP-based and CNN-based networks, respectively. For the pose estimator $P$, we leverage the vision transformer \cite{dosovitskiy2020image} where the output of last layer is modified to an estimated camera pose. The number of feature embeddings in the scene codebook is set to 1024 and the dimension of each feature embedding is set to 512. The dimensions of obtained weights $X$ and $X'$ are the same as the number of feature embeddings in the scene codebook. In term of NeRF in IR-NeRF, we adopt the hierarchical volume sampling strategy \cite{mildenhall2020nerf} to simultaneously optimize `coarse' and `fine' networks to represent scenes. The MLPs in `coarse' and `fine' networks are shared and the dimension of MLPs is set to 360 \cite{meng2021gnerf}. We sample 64 locations along each camera ray in both stratified sampling and inverse transform sampling \cite{mildenhall2020nerf}. The Adam optimizer is adopted to train our IR-NeRF and the mini-batch size is set to 12 for both synthetic and real scenes. We use the Pytorch framework in implementation and employ one NVIDIA RTX 3090ti GPU for both training and inference. 

\renewcommand\arraystretch{1.1}
\begin{table}[t]
    \renewcommand\tabcolsep{3.1pt}
	\begin{center}
	\begin{tabular}{*{7}{c}}
		\toprule
		 \multirow{2}{*}{Scenes} & \multicolumn{2}{c}{PSNR$\uparrow$ }  & \multicolumn{2}{c}{SSIM$\uparrow$ } & \multicolumn{2}{c}{LPIPS$\downarrow$ } \\
		
		\cmidrule(lr){2-3}\cmidrule(lr){4-5}\cmidrule(lr){6-7} 
		
		& \multicolumn{1}{c}{GNeRF} & \multicolumn{1}{c}{\textbf{Ours}} &
		\multicolumn{1}{c}{GNeRF} & \multicolumn{1}{c}{\textbf{Ours}} &
		\multicolumn{1}{c}{GNeRF} & \multicolumn{1}{c}{\textbf{Ours}} \\
		
		\hline
		Chair & 31.30 & \textbf{32.87} & 0.94 & \textbf{0.96} & 0.08 & \textbf{0.07} \\
		Drums & 24.30 & \textbf{25.98} & 0.90 & \textbf{0.91} & 0.13 & \textbf{0.11} \\
		Hotdog & 32.00 & \textbf{33.52} & 0.96 & \textbf{0.97} & 0.07 & \textbf{0.06} \\
		Lego & 28.52 & \textbf{30.07} & 0.91 & \textbf{0.93} & 0.09 & \textbf{0.07} \\
            Mic & 31.07 & \textbf{32.33} & 0.96 & \textbf{0.97} & 0.06 & \textbf{0.04} \\
		Ship & 26.51 & \textbf{27.96} & 0.85 & \textbf{0.87} & 0.21 & \textbf{0.18} \\
		\hline
		\hline
		
		Scan23 & 17.89 & \textbf{19.96} & 0.55 & \textbf{0.59} & 0.54 & \textbf{0.45} \\
		Scan45 & 18.06 & \textbf{20.19} & 0.68 & \textbf{0.73} & 0.48 & \textbf{0.41} \\
		Scan58 & 21.83 & \textbf{24.02} & 0.62 & \textbf{0.67} & 0.67 & \textbf{0.55}\\
		Scan82 & 19.91 & \textbf{21.55} & 0.77 & \textbf{0.85} & 0.33 & \textbf{0.27} \\
		Scan103 & 22.67 & \textbf{24.72} & 0.74 & \textbf{0.82} & 0.44 & \textbf{0.37} \\
		Scan109 & 22.88 & \textbf{25.36} & 0.71 & \textbf{0.75} & 0.54 & \textbf{0.44}  \\
		\bottomrule
	\end{tabular}
   	\end{center}
	\caption{\textbf{Quantitative comparisons of novel view synthesis} on the dataset Synthetic-NeRF and DTU. The proposed IR-NeRF outperforms the state-of-the-art GNeRF consistently in PSNR, SSIM and LPIPS under different synthetic and real scenes. All methods are trained with the same training data and batch size.
	}
	\label{tab_synthetic}

\end{table}

\subsection{Comparisons with the State-of-the-Art}

\paragraph{Novel View Synthesis.}
We compare IR-NeRF with the most related work GNeRF \cite{meng2021gnerf} over different synthetic and real scenes. We did not compare with NeRF$--$ \cite{wang2021nerf}, BARF \cite{lin2021barf}, SCNeRF \cite{jeong2021self} and GARF \cite{chng2022garf} as the four methods require reasonable camera pose initialization and are not applicable to random camera pose initialization. As there is no available pretrained models, we train GNeRF based on its official codes and all methods (including IR-NeRF) are trained with the same training dataset and training setting in experiments. Table~\ref{tab_synthetic} shows experimental results over the same test images as described in section~\ref{dataset}. We can observe that IR-NeRF outperforms the state-of-the-art GNeRF consistently in PSNR, SSIM and LPIPS across all synthetic and real scenes. The superior performance is largely attributed to our proposed implicit pose regularization that allows to refine pose estimator with unposed real images which further improves the robustness of pose estimation for real images.
The quantitative experimental results are well aligned with the qualitative results in Figs.~\ref{visual_compare} where IR-NeRF produces superior multi-view images with less artifacts and finer details.

\paragraph{Camera Pose Estimation.}
We also compare the accuracy of the estimated camera poses of real images as used in NeRF training. The evaluation is performed over the dataset Synthetic-NeRF. For the evaluation metric, we adopt mean camera rotation difference (Rot) and mean translation difference (Trans) that are computed with the toolbox \cite{zhang2018tutorial} on the training set. As Table \ref{tab_pose} shows, IR-NeRF outperforms GNeRF clearly and consistently across all evaluated scenes. The superior estimation accuracy is largely attributed to our designed implicit pose regularization. The robustness of camera pose estimation for real images can be improved with this pose regularization, further leading to superior joint refinement of camera poses and NeRF without falling into local minima.

\begin{table}[t]
\renewcommand\tabcolsep{9pt}
	\begin{center} 
	\begin{tabular}{*{5}{c}}
		\toprule
		\multirow{2}{*}{Scenes} & \multicolumn{2}{c}{GNeRF \cite{meng2021gnerf}} & \multicolumn{2}{c}{IR-NeRF} \\
		
		\cmidrule(lr){2-3}\cmidrule(lr){4-5}
		
		& \multicolumn{1}{c}{Rot($^{\circ}$)$\downarrow$} & \multicolumn{1}{c}{Trans$\downarrow$ } & \multicolumn{1}{c}{Rot($^{\circ}$)$\downarrow$ } & \multicolumn{1}{c}{Trans$\downarrow$ } \\
		
		\hline
		Chair & 0.363 & 0.018 & \textbf{0.251} & \textbf{0.013}  \\
		Drums & 0.204 & 0.010 & \textbf{0.185} & \textbf{0.008}  \\
		Hotdog & 2.349 & 0.122 & \textbf{1.932} & \textbf{0.098} \\
		Lego & 0.430 & 0.023 & \textbf{0.371} & \textbf{0.015} \\
	    Mic & 1.865 & 0.031 & \textbf{1.598} & \textbf{0.019}  \\
		Ship & 3.721 & 0.176 & \textbf{3.253} & \textbf{0.125} \\
		\bottomrule
	\end{tabular}
 	\end{center}
	\caption{\textbf{Quantitative comparisons of the accuracy of camera pose estimation} (on Synthetic-NeRF): Rot and Trans represent mean camera rotation differences and mean camera translation differences, respectively. 
	IR-NeRF outperforms the state-of-the-art GNeRF consistently in Rot and Trans in all studied scenes. }
	\label{tab_pose}
\end{table}

\begin{figure*}[t]
\begin{center}
\includegraphics[width=1\textwidth]{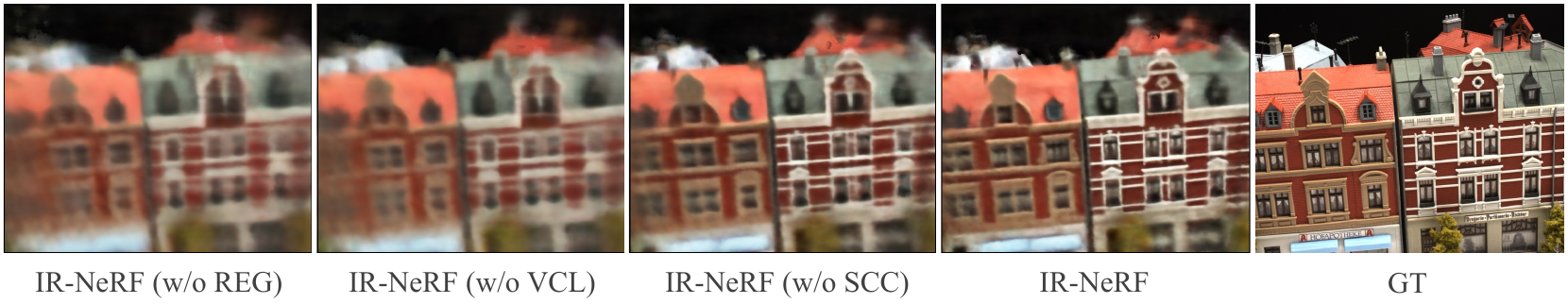}
\end{center}
\caption{
\textbf{Qualitative ablation studies of the proposed IR-NeRF:} IR-NeRF and its variants (including IR-NeRF (w/o REG), IR-NeRF (w/o VCL), and IR-NeRF (w/o SCC) that remove the proposed implicit pose regularization, view consistency loss, and scene codebook construction, respectively) are trained on the scan15 of DTU dataset. \textbf{Zoom in for best view.}}
\label{visual_abla}
\end{figure*}

\subsection{Ablation Studies}

\paragraph{Effect of Implicit Pose Regularization .} 
We examine the contribution of our proposed implicit pose regularization. As Table \ref{ablation} shows, we train the model \textit{IR-NeRF (w/o REG)} by removing the implicit pose regularization from the \textit{IR-NeRF}. The IR-NeRF (w/o REG) does not involve the two
key components so it can be regarded as a baseline that
trains the pose estimator in the similar way as GNeRF. It can be seen that \textit{IR-NeRF (w/o REG)} degrades PSNR, SSIM and LPIPS significantly as compared with \textit{IR-NeRF}, indicating that the proposed implicit pose regularization can effectively improve the robustness of pose estimation and further achieve superior novel view synthesis for IR-NeRF. The effectiveness of the proposed implicit pose regularization can be observed in Fig.~\ref{visual_abla} as well where the model \textit{IR-NeRF} can produce clearer visual results than the model \textit{IR-NeRF (w/o REG)}.

\paragraph{Effect of Scene Codebook Construction.} 
To examine the effectiveness of the designed scene codebook construction, we study how it affects view synthesis in PSNR, SSIM and LPIPS. As shown in Table \ref{ablation}, we train \textit{IR-NeRF (w/o SCC)} that removes the designed scene codebook construction from the complete model \textit{IR-NeRF}. Quantitative experiments show that \textit{IR-NeRF} performs clearly better than the \textit{IR-NeRF (w/o SCC)} in novel view synthesis, demonstrating the effectiveness of pre-learning a scene codebook for subsequent pose-guided view reconstruction. The experimental results are also well aligned with qualitative experimental results in Fig.~\ref{visual_abla} where \textit{IR-NeRF} with the designed scene codebook construction synthesizes novel views with less artifacts and finer details compared with the synthesis by \textit{IR-NeRF (w/o SCC)}.

\renewcommand\arraystretch{1.2}
\small
\begin{table}[t]
\renewcommand\tabcolsep{6.0pt}
\begin{center}
    
\begin{tabular}{l||ccc} 
\hline
& 
\multicolumn{3}{c}{\textbf{Evaluation Metrics}}
\\
\cline{2-4}
\multirow{-2}{*}{\textbf{Models}} 
& PSNR $\uparrow$ 
& SSIM $\uparrow$
& LPIPS $\downarrow$ 
\\\hline
\textbf{IR-NeRF (w/o REG)} & 17.05 & 0.53 & 0.65  \\

\textbf{IR-NeRF (w/o SCC)} & 18.23 & 0.55 & 0.54  \\

\textbf{IR-NeRF (w/o VCL)} & 17.38 & 0.54 & 0.64  \\
\hline
\textbf{IR-NeRF} & \textbf{19.88} & \textbf{0.59} & \textbf{0.47} \\

\hline
\end{tabular}
\end{center} 
\caption{
\textbf{Ablation studies of the proposed IR-NeRF} on the scene `scan23' of DTU. \textit{IR-NeRF (w/o REG)} removes the implicit pose regularization (REG) from IR-NeRF, which is equivalent to baseline. \textit{IR-NeRF (w/o SCC)} removes the scene codebook construction (SCC), where input image is naively encoded to latent features. \textit{IR-NeRF (w/o VCL)} removes the view consistency loss (VCL) in the pose-guided view reconstruction, where a reconstruction loss is introduced between pose-guided reconstructed image and real image. All models are trained with same training settings.
}
\label{ablation}

\end{table}

\paragraph{Effect of View Consistency Loss.}
We further examine the view consistency loss in the pose-guided view reconstruction by comparing the model \textit{IR-NeRF (w/o VCL)} and \textit{IR-NeRF}. As Table \ref{ablation} shows, adopting view consistency loss improves PSNR, SSIM and LPIPS consistently, indicating the effectiveness of our designed view consistency loss in pose-guided view reconstruction. The quantitative results are well aligned with the qualitative experiments in Fig.~\ref{visual_abla} as well.

\begin{figure}[t]
\begin{center}\includegraphics[width=1\linewidth]{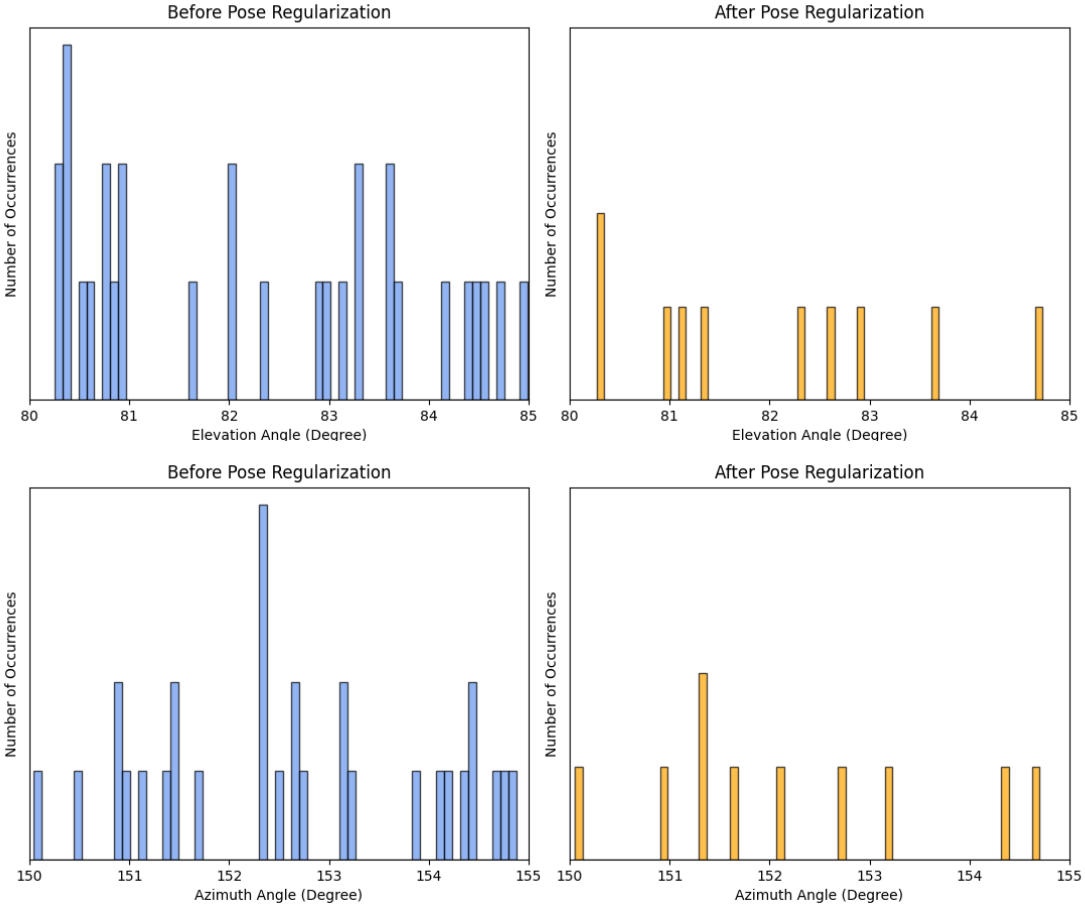}
\end{center}
\caption{
\textbf{
Visualization of estimated out-of-distribution camera poses}: Camera poses are estimated from real images of the DTU dataset.
We focus on the elevation and azimuth angles of the estimated camera poses, which are two key parameters in camera pose distribution. The range ([150$^{\circ}$, 155$^{\circ}$]) of the elevation angle and the range ([80$^{\circ}$, 85$^{\circ}$]) of the azimuth angle are out of the camera pose distribution for the DTU scenes. The y-axis represents the number of occurrences. It can be observed that after applying the proposed implicit pose regularization, the estimated out-of-distribution camera poses (in azimuth angle and elevation angle) are significantly reduced.
}
\label{visualization}
\end{figure}

\subsection{Visualization}

We visualize out-of-distribution camera poses estimated before and after the proposed implicit pose regularization by using histograms. As Fig.~\ref{visualization} shows, much less out-of-distribution camera poses are predicted after applying the proposed implicit pose regularization. This shows that the proposed implicit pose regularization can effectively refine the pose estimation and improve the robustness of pose estimation for real images, which greatly helps to mitigate local minima in the subsequent joint refinement of camera poses and NeRF.




\section{Limitation}

Although the proposed IR-NeRF achieves superior NeRF training by implicit pose regularization as compared with state-of-the-art GNeRF, it still has one major limitation. Specifically, the training process of IR-NeRF includes coarse NeRF learning, coarse camera pose estimation, and joint refinement of camera poses and NeRF, which requires a long training time. Moving forward, we will focus on pose-free NeRF training at much higher speed. The training speed could potentially be improved by introducing more efficient representation, such as triplane and tensor decomposition.

\section{Conclusion}

This paper presents IR-NeRF, a pose-free NeRF with implicit pose regularization that promotes the robustness of pose estimation for real images, thus preventing the joint refinement of NeRF and predicted camera poses from falling into local minima. 
Given a set of multi-view images of a scene, we construct a scene codebook to encode scene features and capture scene-specific pose distribution as priors. In addition, we design pose-guided view reconstruction with view consistency loss which refines pose estimation for real images with the scene priors based on the rationale that a real image can be reconstructed well from the learned scene codebook only when its estimated camera pose lies within the scene-specific pose distribution. Extensive experiments over synthetic and real scenes demonstrate the superiority of IR-NeRF. 

\section{Acknowledgements}

This project is funded by the Ministry of Education Singapore, under the Tier-2 project scheme with a project number MOE-T2EP20220-0003.
Fangneng Zhan is supported by the ERC Consolidator Grant 4DReply (770784).

{\small
\bibliographystyle{ieee_fullname}
\bibliography{egbib}
}

\end{document}